\documentclass[letterpaper, 10 pt, conference]{ieeeconf}  

\usepackage[firstpage=true]{background}
\newcommand\copyrighttext{%
\parbox{\textwidth}{
\footnotesize
\textbf{Accepted final version.} IEEE International Conference on Robotics and Automation (ICRA), Xi'an, China, to appear June 2021.
}
}

\SetBgContents{\copyrighttext}
\SetBgScale{1}
\SetBgColor{black}
\SetBgAngle{0}
\SetBgOpacity{1}
\SetBgPosition{current page.north}
\SetBgVshift{-0.8cm}

\usepackage[bottom]{footmisc}

\IEEEoverridecommandlockouts                              

\usepackage[utf8]{inputenc}
\usepackage[T1]{fontenc}
\usepackage[hidelinks]{hyperref}
\usepackage{graphicx}
\usepackage{tabularx}
\usepackage{multirow}
\usepackage{amsmath,amssymb}
\usepackage{xspace}
\usepackage{balance}
\usepackage[per-mode=symbol,binary-units=true,range-units=single,range-phrase=-,detect-weight=true,detect-family=true]{siunitx}
\usepackage{booktabs}
\usepackage{paralist}
\usepackage{subcaption}
\usepackage[font=footnotesize]{caption}
\usepackage{url}
\usepackage{tikz}
\usepackage{pgfplots}
\usetikzlibrary{fit,shapes.callouts,shapes.geometric,backgrounds,positioning,arrows.meta}
\usepackage[ruled,vlined]{algorithm2e}

\newcommand{\ie}{i.e.,\ }

\newcommand{\etal}{\xspace{}et al.\xspace}

\newcommand{\reffig}[1]{Fig.~\ref{#1}}
\newcommand{\reftab}[1]{Tab.~\ref{#1}}
\newcommand{\refsec}[1]{Sec.~\ref{#1}}
\title{\LARGE \bf
Search-based Planning of Dynamic MAV Trajectories Using Local Multiresolution State Lattices
}

\author{Daniel Schleich and Sven Behnke
\thanks{All authors are with the Autonomous Intelligent Systems group, 
		University of Bonn, Germany;
        {\tt\small schleich@ais.uni-bonn.de}}
\thanks{ This work has been funded by the German Federal Ministry of Education and Research (BMBF) in the project ``Kompetenzzentrum: Aufbau des Deutschen Rettungsrobotik-Zentrums (A-DRZ)''.}
}

\begin{document}

\maketitle
\thispagestyle{empty}
\pagestyle{empty}

\begin{abstract}
Search-based methods that use motion primitives can incorporate the system's dynamics into the planning and thus generate dynamically feasible MAV trajectories that are globally optimal.
However, searching high-dimensional state lattices is computationally expensive.
Local multiresolution is a commonly used method to accelerate spatial path planning.
While paths within the vicinity of the robot are represented at high resolution, the representation gets coarser for more distant parts.
In this work, we apply the concept of local multiresolution to high-dimensional state lattices that include velocities and accelerations.
Experiments show that our proposed approach significantly reduces planning times.
Thus, it increases the applicability to large dynamic environments, where frequent replanning is necessary.
\end{abstract}

\section{Introduction}
\label{sec:introduction}
The application of micro aerial vehicles (MAVs) to surveillance, inspection, and search \& rescue  tasks has gained increasing popularity in recent years.
Generating fast, dynamically feasible MAV trajectories for such scenarios is a challenging task, since one usually has to deal with large, initially unknown environments.
Fast trajectory replanning is necessary to avoid dynamic or previously unknown obstacles.
Many existing methods generate trajectories in a two-stage approach:
First, a 3D position-only obstacle-free path is planned using search-based or sampling-based methods~\cite{hart1968formal, lavalle1998rapidly}.
Afterwards, this path is refined to a high-dimensional trajectory including velocities.
Refinement strategies include quadratic programming~\cite{richter2016polynomial}, B-Spline path  planning~\cite{koyuncu2008probabilistic} and gradient-based optimization methods~\cite{kalakrishnan2011stomp} as used in~\cite{nieuwenhuisen2019search}.
These approaches only generate locally optimal trajectories.
However, incorporating the system's dynamics into the planning process allows to generate globally optimal trajectories.
Liu \etal~\cite{liu2017search} propose a search-based method to directly plan second- and third-order MAV trajectories.
They unroll motion primitives to generate a state lattice graph, from which they extract trajectories using A* with a heuristic based on the solution of a Linear Quadratic Minimum Time problem.
Since velocities are explicitly considered during planning, their approach can be extended to generate trajectories through doors that are narrower than the MAV diameter~\cite{liu2018search}.
However, searching high-dimensional state spaces is computationally expensive.
Thus, such approaches either lack the capability for fast replanning or are restricted to small or low-resolutional environment representations.
A common approach for reducing the state space size for spatial path planning is local multiresolution~\cite{behnke2003local}.
Here, only the vicinity of the robot is represented at high resolution, while the resolution decreases with increasing distance from the robot.
In this work, we apply the concept of local multiresolution to the state lattice graph of Liu \etal~\cite{liu2017search}.
Thus, we are able to significantly reduce planning times.
This increases the applicability of the approach to large dynamic environments, where frequent replanning is necessary.
In summary, the main contributions of this paper are:
\begin{itemize}
 \item The introduction of local multiresolutional state lattices,
 \item an expansion scheme for the A* algorithm addressing the issues of multiresolutional state representations, and
 \item a search heuristic based on the solution of 1D sub-problems.
\end{itemize}

\begin{figure}[t]
\begin{center}
 \includegraphics[width=0.4\textwidth]{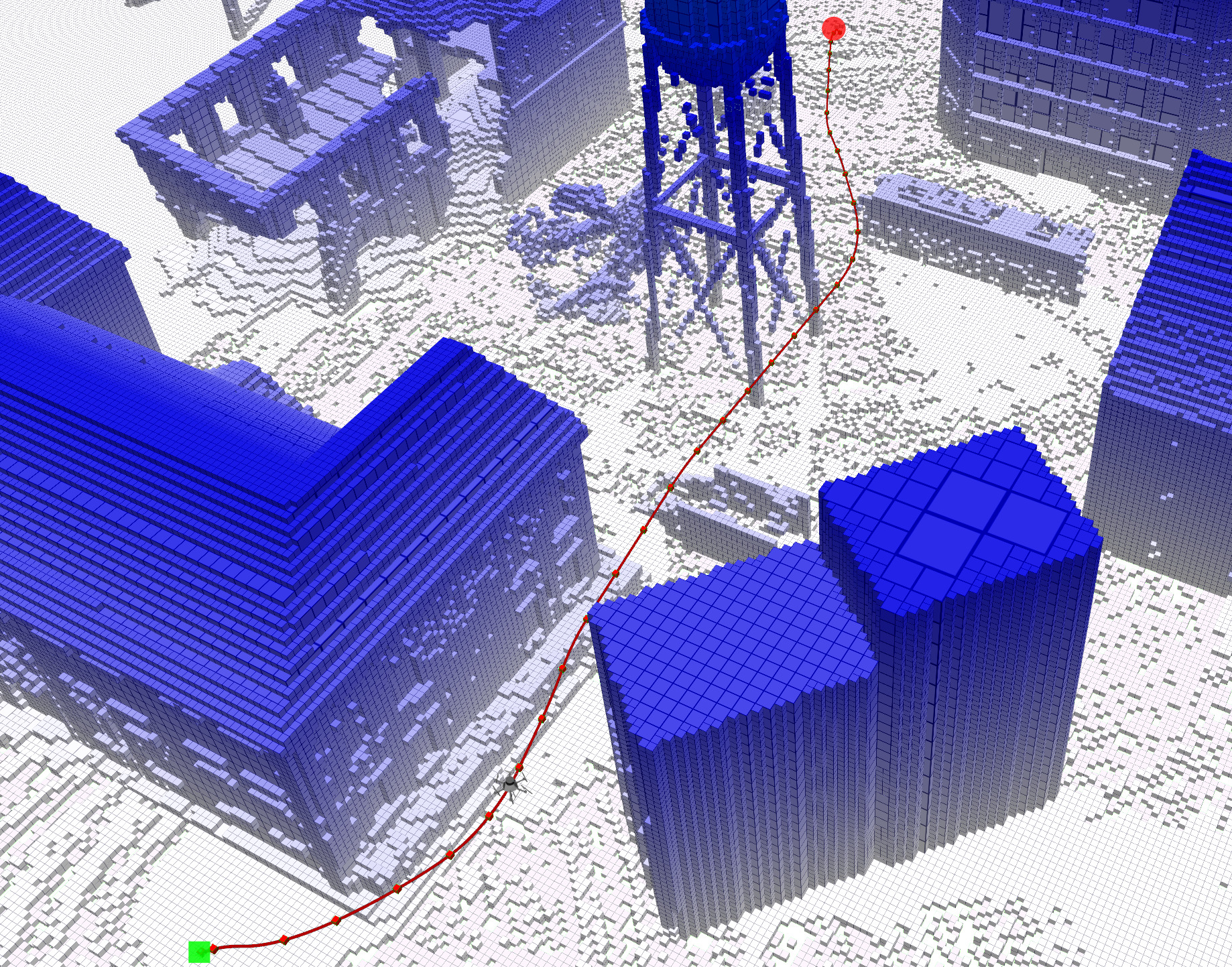}
 \end{center}
 \caption{ The MAV following a trajectory in a simulated outdoor environment. The start is at the red circle and the goal is marked by the green square. The trajectory is replanned online at \SI{1}{\hertz}. Red cubes mark the start positions of each replanning step. }
 \vspace{-0.5cm}
 \label{fig:teaser}
\end{figure}
 
\section{Related Work}
\label{sec:related_work}
Multiresolutional approaches are commonly used to accelerate planning.
Behnke~\cite{behnke2003local} proposed A*-based path planning on robot-centered local multiresolution grids, where the spatial resolution decreases with increasing distance from the robot.
Nieuwenhuisen \etal~\cite{nieuwenhuisen2014hierarchical} extend this approach to plan 3D trajectories for MAVs.
In~\cite{nieuwenhuisen2016local}, local multiresolution in time is used to enable fast reoptimization of an initial path from a grid-based planner.
Du \etal~\cite{du2020multi} perform multiple A* searches simultaneously on grids with different resolutions while sharing states that lie on multiple resolution grids.

Multiresolution has also been used in combination with state lattices.
Gonz\'alez-Sieira \etal~\cite{gonzalez2016adaptive} choose the resolution level based on the complexity of the local environment, \ie on the distance from obstacles.
They apply different sets of motion primitives for each level.
In~\cite{gonzalez2019graduated}, the same authors group motion primitives in categories.
When expanding a state, the longest collision-free motion from each category is applied.
Pivtoraiko \etal~\cite{pivtoraiko2008differentially} use a uniform 2D grid but apply different state transitions for start and goal area than for the other areas.
This framework is extended by Andersson \etal~\cite{andersson2018receding}.
They propose a receding-horizon method, where the first half of the planning time is used for planning in a high-dimensional space.
The search is continued afterwards with lower dimension and a reduced action set.
All of the above state lattice methods use different action sets dependent on the current representation level.
However, all levels share the same spatial resolutions.
In contrast to those methods, our approach reduces the spatial resolution with increasing distance from the current MAV position.

Likhachev \etal~\cite{likhachev2009planning} propose multi-resolutional state lattices to plan trajectories for autonomous ground vehicles.
The action set of the lower-resolution level is a strict subset generated from the higher-level action set by choosing only actions whose end states lie on the low-dimensional grid.
The state representation used by Likhachev \etal only includes movement directions, but their approach is expanded by Rufli \etal~\cite{rufli2009smooth} to also consider velocities.
Petereit \etal~\cite{petereit2013mobile} use two representation levels with different dimensionality.
The representation level of a state is chosen based on the time it takes to reach this state.
Similar to the method of Likhachev \etal, transitions from higher to lower representation levels are only possible for states that match the resolution of the lower level.

Our method uses the same dimensionality for all representation levels but reduces spatial and velocity resolutions for higher levels.
In contrast to the approaches mentioned above, we do not rely on a fixed precomputed set of motion primitives.
For transitions between different representation levels, we adjust the motion primitives such that the resulting states match the resolution of the target level.
 
\section{Method}
\label{sec:method}
Our trajectory planning method is based on the framework of Liu \etal~\cite{liu2017search}.
In this work, we restrict ourselves to second-order systems and leave the extensions to third-order systems for future work.
Thus, we model the MAV state as a $6$-tuple $s=(\mathbf{p}, \mathbf{v}) \in \mathbb R^6$ consisting of a 3D position $\mathbf{p}$ and velocity $\mathbf{v}$.
Mind, that we do not explicitly model the MAV yaw, since it does not influence the system's dynamics.
Instead, we choose the yaw in a post-processing step based on the flight direction of the planned trajectory.

\begin{figure}
\begin{center}
\hspace*{-0.3em}a)\hspace{0.4em}\mbox{\includegraphics[width=0.2\textwidth]{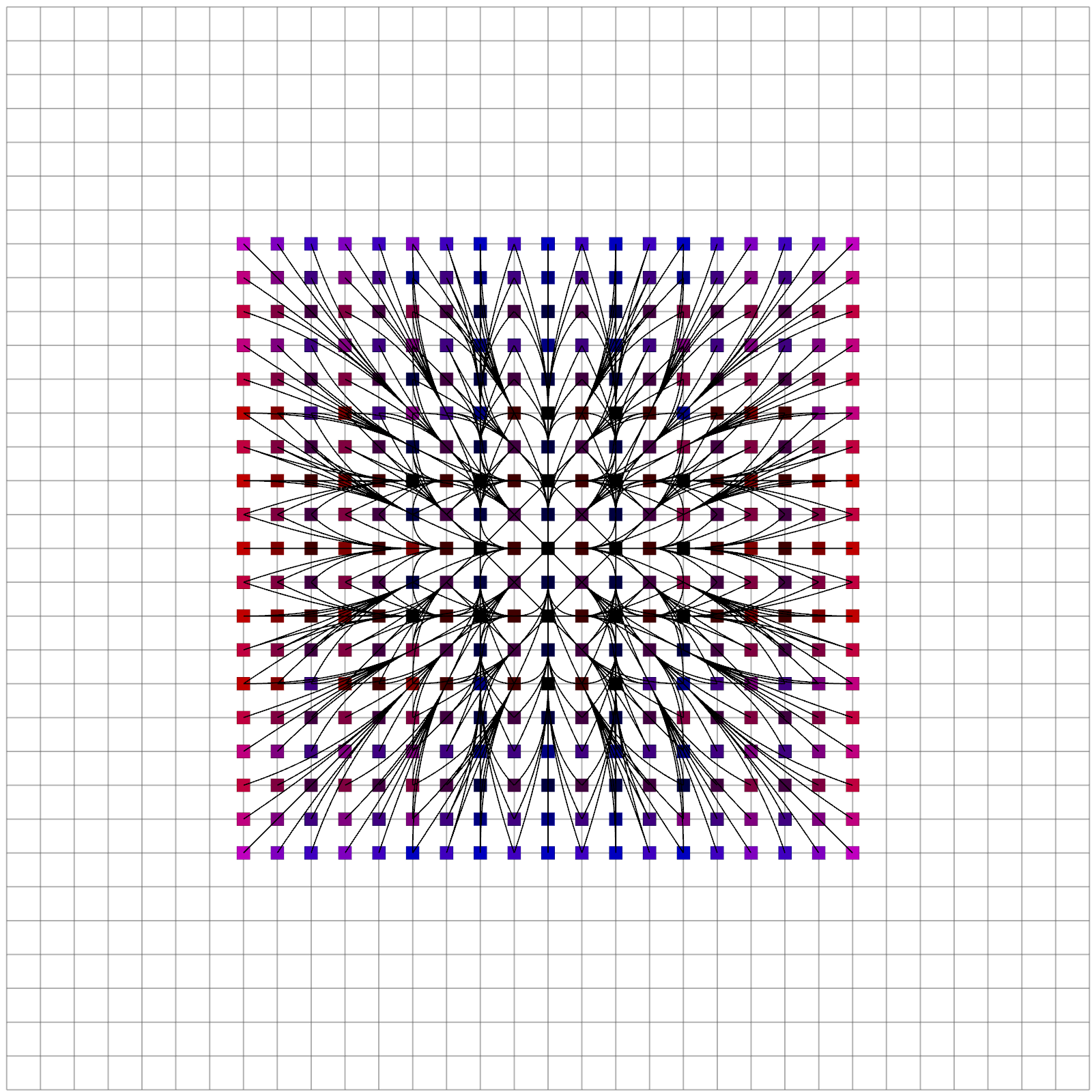}}
  \hspace{0.4em}
b)\hspace{0.4em}\mbox{\includegraphics[width=0.2\textwidth]{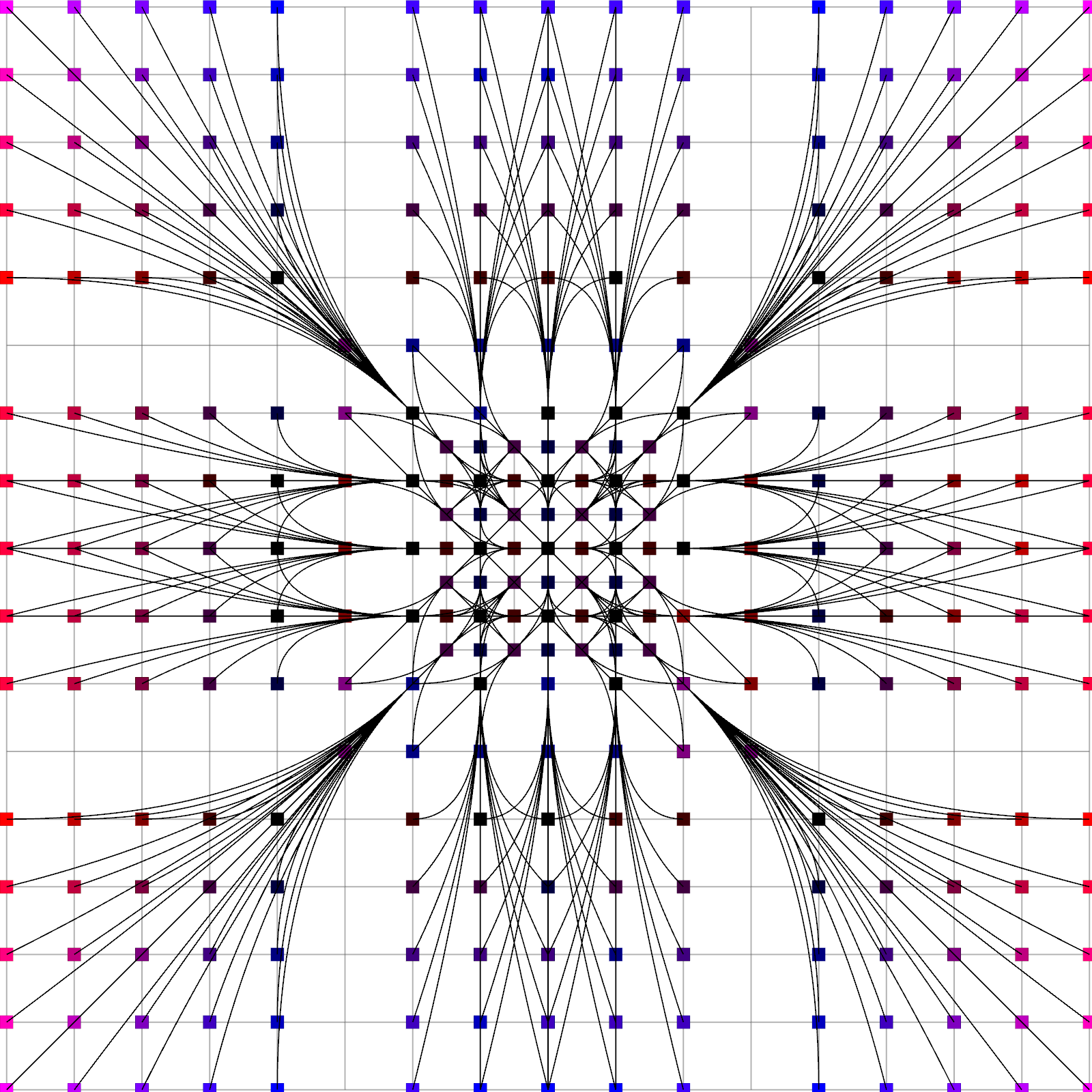}}
 \end{center}
 \caption{Top-down view of 2D state lattice graphs. a) Uniform. b) Local multiresolution. The spatial position of nodes is fixed to the corners of a multiresolution grid with high resolution at the center, \ie the current MAV position, and coarser resolution for more distant areas. Bright red represents high absolute velocity along the x-axis, bright blue represents high velocity along the y-axis. }
 \label{fig:state_lattices}
  \vspace{-1em}
\end{figure}

The state space is discretized by unrolling motion primitives from the initial MAV state $s_0=(\mathbf{p_0}, \mathbf{v_0})$, resulting in a state lattice graph $\mathcal G(\mathcal S, \mathcal E)$.
Here, $\mathcal S$ denotes the set of discretized MAV states, and $\mathcal E$ is the set of motion primitives connecting the states.
Each motion primitive is generated by applying a constant acceleration $\mathbf{u}$ from a discrete control set $\mathcal U_M \subset \mathbb R^3$ over a short time interval $\tau$.
Thus, the motion primitive $F_{\mathbf{u},s}$ connecting the states $s$ to $s':=F_{\mathbf{u},s}(\tau)$ can be expressed as a time-parameterized polynomial
\begin{equation}
F_{\mathbf{u},s}(t) = \begin{pmatrix}\mathbf{p}+ t \mathbf{v} + \frac{t^2}{2} \mathbf{u}\\\mathbf{v} + t \mathbf{u}\end{pmatrix},\,\text{ for } t\in[0,\tau].
\label{eq:primitive}
\end{equation}
An example for the resulting state lattice graph is depicted in~\reffig{fig:state_lattices}\,a.

The costs for a motion primitive are defined as the weighted sum of control effort and primitive duration, \ie
\begin{equation}
C(F_{\mathbf{u},s}) = ||\mathbf{u}||^2_2\tau + \rho\tau.
\label{eq:primitive_cost}
\end{equation}

The optimal trajectory can be planned in the state lattice graph $\mathcal G$ by applying graph search methods like A*.
For more details, we refer to~\cite{liu2017search}.

In the following, we introduce multiple concepts to reduce planning times.
In~\refsec{sec:mres_lattice}, we reduce the state space size by applying the concept of local multiresolution to the state lattice graph $\mathcal G$.
The idea is to restrict the state positions to the corners of a MAV-centered local multiresolution grid, as shown in~\reffig{fig:state_lattices}\,b.
Additionally, we reduce the number of discrete velocities for states whose positions are far from the current MAV position.

In~\refsec{sec:heuristic}, we propose a search heuristic based on the solution of 1D sub-problems.
To avoid overshooting the goal position, we introduce special goal action sequences in~\refsec{sec:goal_actions}.
Finally,~\refsec{sec:level_astar} details an expansion scheme for the A* algorithm, specially designed to overcome issues of multiresolutional state representations.

\subsection{Local Multiresolutional State Lattices}
\label{sec:mres_lattice}
We define multiple levels of state discretization, dependent on the distance to the initial MAV position $\mathbf{p_0}$.
\textit{Level-1} covers all states whose positions are close to $\mathbf{p_0}$.
The resolution represents the smallest possible position and velocity changes $\Delta^{(1)}_p$ and $\Delta^{(1)}_v$ of a motion primitive.
They depend on the primitive duration $\tau$ and the minimal non-zero acceleration command $u_\text{min}$, and they can be obtained from \eqref{eq:primitive}:
\begin{equation}
\Delta^{(1)}_p = \frac{1}{2}\tau^2 u_\text{min}\,\text{, } \Delta^{(1)}_v = \tau u_\text{min}.
\label{eq:resolutions}
\end{equation}
The spatial resolutions of the higher levels are obtained by halving the resolution of the next lower level.
For \textit{Level}-$i$, we get $\Delta^{(i)}_p:=2^{i-1}\Delta^{(1)}_p$.
If we choose $\tau < 1$, the velocity resolution is much coarser than the spatial resolution.
Therefore, we only halve the velocity resolution after every second resolution level, \ie we define
\begin{equation}
 \Delta^{(2)}_v := \Delta^{(1)}_v,\text{ and } \Delta^{(3)}_v := \Delta^{(4)}_v := 2\Delta^{(1)}_v.
\end{equation}
Each resolution level covers an eight times larger volume than the next lower level.
They are embedded into each other such that the center of a higher level is replaced by the next lower level.
\reffig{fig:state_lattices}\,b shows an example with two levels.

When generating the state lattice graph by unrolling motion primitives, we have to ensure that the target states of the motion primitives match the discretization of the corresponding resolution level.
This is done by carefully defining control sets $\mathcal U^{(i)}$ and time steps $\tau^{(i)}$ for each level.
Mind that we define control sets independently for each spatial dimension.
With respect to the choice of the time steps, there are two different approaches, which we explain in the following and evaluate against each other in~\refsec{sec:evaluation}.

\paragraph{Fixed time steps per level}
From \eqref{eq:resolutions}, it follows that the velocity resolution is halved when doubling the time step, while the spatial resolution is reduced by a factor of four.
Thus, by choosing $\tau^{(i)} := 2^{i-1}\tau^{(1)}$, we ensure that position and velocity changes are multiples of the \textit{Level-}$i$ resolutions.
If we additionally halve the acceleration commands, the velocity resolution does not change.
Thus, for each command $u^{(1)}\in\mathcal U^{(1)}$, we define the corresponding higher-level commands as
\begin{equation}
 u^{(2)} := u^{(3)} := \frac{1}{2}u^{(1)}\,\text{ and } u^{(4)} := \frac{1}{4}u^{(1)}.
\end{equation}
Halving the commands while doubling durations significantly reduces maneuverability, however.
We mitigate this effect by adding two special actions to each level: decelerate to zero velocity and accelerate to the maximum allowed velocity in current flight direction.

\begin{figure}
\begin{center}
a)\hspace{0.5em}\mbox{\includegraphics[width=0.149\textwidth]{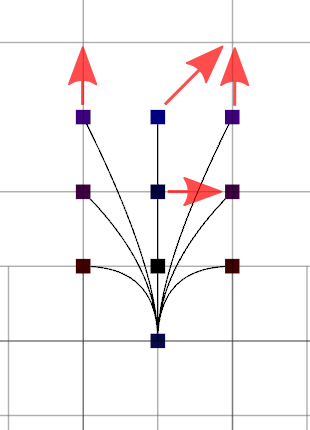}}
  \hspace{0.7em}
b)\hspace{0.5em}\mbox{\includegraphics[width=0.149\textwidth]{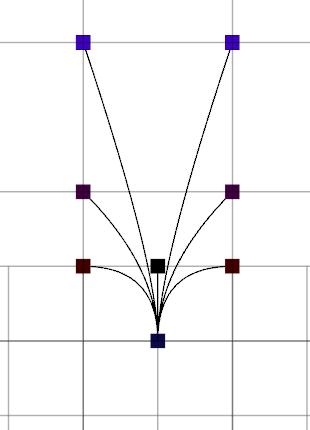}}
 \end{center}
  \vspace{-0.2em}
 \caption{Adjusting motion primitives to the local multiresolution grid. First, the lower-level motion primitives are unrolled (a). Then, the closest grid corners to the primitive end positions are determined (red arrows). Finally, primitives ending at those grid corners are generated (b). }
 \label{fig:primitive_adjustment}
  \vspace{-0.8em}
\end{figure}

The above method enforces that end states of all motion primitives starting in \textit{Level-}$i$ also lie on the \textit{Level-}$i$ grid.
However, level transitions have to be specially addressed (see~\reffig{fig:primitive_adjustment}):
We predict the end state of a motion primitive, determine the closest position $\mathbf{p_g}$ on the target resolution grid, and choose the acceleration command $\mathbf{u}$ such that the primitive ends at $\mathbf{p_g}$:
\begin{equation}
\mathbf{u}=2\frac{\mathbf{p_g}-\mathbf{p}-\tau^{(i)}\mathbf{v}}{(\tau^{(i)})^2},
\end{equation}
where $(\mathbf{p}, \mathbf{v})$ denotes the start state of the motion primitive.
Using $\Delta_v = \tau^{(i)}\mathbf u$, it follows that the resulting velocity change always is a multiple of the \textit{Level-1} velocity resolution, but for higher levels, the velocity might not match the coarser resolution.
Intermediately, we allow this offset for the end states of the adjusted primitives
and correct it only when generating their successors.

\paragraph{Variable time steps per level}
Position changes do not only depend on $\tau$ and $u_\text{min}$ but also on the initial velocity.
Using large time steps for high velocities results in a much lower spatial resolution.
Thus, many motion primitives with high initial velocities become invalid in the presence of obstacles.
Valid trajectories can still be found by choosing lower velocities.
However, this increases the trajectory costs, which leads to a significant increase of node expansions during the search (compare~\refsec{sec:eval_astar}).
Thus, it might be a good idea to choose the time steps dependent on the current velocity.
For each command $u$ and initial velocity $v$, we choose the smallest time step $\tau^{(i)}\in\{2^k\tau^{(1)},\,k\in\mathbb N_0\}$, such that the position change $\tau^{(i)}v + \frac{1}{2}(\tau^{(i)})^2 u$ is larger than the spatial resolution $\Delta_p^{(i)}$ of the current level.
If the position does not change, we set $\tau^{(i)}=\tau^{(1)}$.
Mind that we use the same command set for each resolution level.
Additionally, we adjust the motion primitive end points to the multiresolution grid as described above.

\subsection{1D Heuristic}
\label{sec:heuristic}
Liu \etal~\cite{liu2017search} use a heuristic based on the solution of a Linear Quadratic Minimum Time problem.
Their heuristic considers constraints on the maximum velocity but assumes continuous instead of piecewise-constant control commands.
Furthermore, the commands are only implicitly bounded by optimizing the control effort and thus might violate constraints on the maximum acceleration.

We propose a heuristic based on precomputing the actual costs for a 1D problem.
To reduce the size of the look-up table, we assume that all goal states have zero velocity.
For each pair of signed distance to the goal position and start velocity, we precompute the costs of the optimal 1D trajectory using the \textit{Level-1} resolutions.

The costs of a trajectory $\mathbf{u}_{0:N-1}$ of length $N$ and duration $T=\sum_{k=0}^{N-1}\tau_k$ can be approximated by
\begin{equation}
\begin{split} 
& C(\mathbf{u}_{0:N-1}) = \rho T + \sum\limits_{k=0}^{N-1} ||\mathbf{u}_k||_2^2 \tau_k \\
&= \rho T + \sum\limits_{k=0}^{N-1} {(u_k)}_x^2 \tau_k + \sum\limits_{k=0}^{N-1} {(u_k)}_y^2 \tau_k + \sum\limits_{k=0}^{N-1} {(u_k)}_z^2 \tau_k \\
&>= \rho T + c_x + c_y + c_z,
\end{split}
\end{equation}
where $c_x, c_y, c_z$ are lower bounds on the control efforts along the individual dimensions.

During search, we look up the times $T^\text{1D}_x, T^\text{1D}_y, T^\text{1D}_z$ and control efforts $c^\text{1D}_x, c^\text{1D}_y, c^\text{1D}_z$ of the corresponding 1D sub-problems.
The total time $T$ of the 3D trajectory is given by $T=\max\{T^\text{1D}_x, T^\text{1D}_y, T^\text{1D}_z\}$.
Without loss of generality, let $T=T^\text{1D}_x$.
Thus, we can set $c_x=c^\text{1D}_x$.
However, $c^\text{1D}_y$ and $c^\text{1D}_z$ might overestimate $c_y$ and $c_z$ if the 1D trajectories have lower durations than the 3D trajectory.
Therefore, we choose $c_y$ and $c_z$ dependent on the current flight direction along the corresponding dimensions:
\begin{itemize}
 \item When flying towards the goal, we choose the control costs that correspond to a full stop.
 \item If the velocity is zero, we choose the control costs that correspond to applying minimal acceleration $u_\text{min}$ followed by deceleration $-u_\text{min}$.
 \item When flying away from the goal, the MAV has to stop, accelerate towards the
goal, and decelerate again. Thus, the resulting control cost is the sum of the above cases.
\end{itemize}
Note that the 1D heuristic is admissible but not consistent:
If the dimension, for which the maximum time is achieved, changes, the control costs of a different 1D sub-problem will be used.
Thus, the decrease of the estimated 3D control costs might be larger than the costs of the applied action.

\subsection{Goal Actions}
\label{sec:goal_actions}
When increasing the motion primitive duration for higher-resolution levels, we might overshoot the goal position frequently during search.
Thus, when expanding a state $s$, we check whether the goal is reachable:
Let $s$ be represented within \textit{Level}-$i$.
We determine lower and upper bounds $p_\text{min}, p_\text{max} \in \mathbb R^3$ of the area that the MAV can reach within the next time step.
Here, we assume a possible acceleration of $\pm u_\text{max}$ for a duration of $2 ^{i-1}\tau^{(1)}$ along each dimension.
If the goal position lies within $[p_\text{min}$, $p_\text{max}]$, we check whether it can be reached from $s$ using a sequence of \textit{Level-1} motion primitives with a total duration of at most $2 ^{i-1}\tau^{(1)}$.
If so, we add the goal state as a neighbor of $s$, connected by the resulting motion primitive sequence.
As for the heuristic, we do not consider any obstacles and assume that the goal state has zero velocity.
Thus, the sequences of \mbox{\textit{Level-1}} motion primitives can be efficiently precomputed for 1D sub-problems and looked up based on the current velocity and distance to the goal.

\subsection{Level-Based Expansion Scheme}
\label{sec:level_astar}
The state space size is significantly reduced when using local multiresolution.
However, experiments show that the A* algorithm tends to expand much more states when applied to multiresolutional state lattices compared to uniform lattices.
We found that a reason for this behavior is the fact that the A* algorithm expands all states whose estimated costs are lower than the costs of the optimal solution (see~\refsec{sec:eval_astar}).
Thus, we adapt the A* algorithm such that it does not always expand the state with the lowest estimated costs, \ie lowest $f$-value, but also might expand states with higher values.
This might result in sub-optimal solutions, which we accept since the multiresolutional lattice representation already introduces sub-optimality.

For each resolution level, we use a separate priority queue.
To determine which node to expand, we compare the state with lowest $f$-value from each level.
A state is considered an expansion candidate if the difference between its $f$-value and the globally minimal value is at most the cost of one step in the corresponding resolution level.
From all candidates, we expand the one with the lowest heuristic value.
We stop the search as soon as the goal state is added to the \textit{OPEN}-list, instead of waiting until the goal state is expanded.
This further reduces the runtime at the cost of possible suboptimality.
 
\section{Evaluation}
\label{sec:evaluation}
\begin{figure}
a)\hspace{0.2em}\mbox{\includegraphics[width=0.25\textwidth]{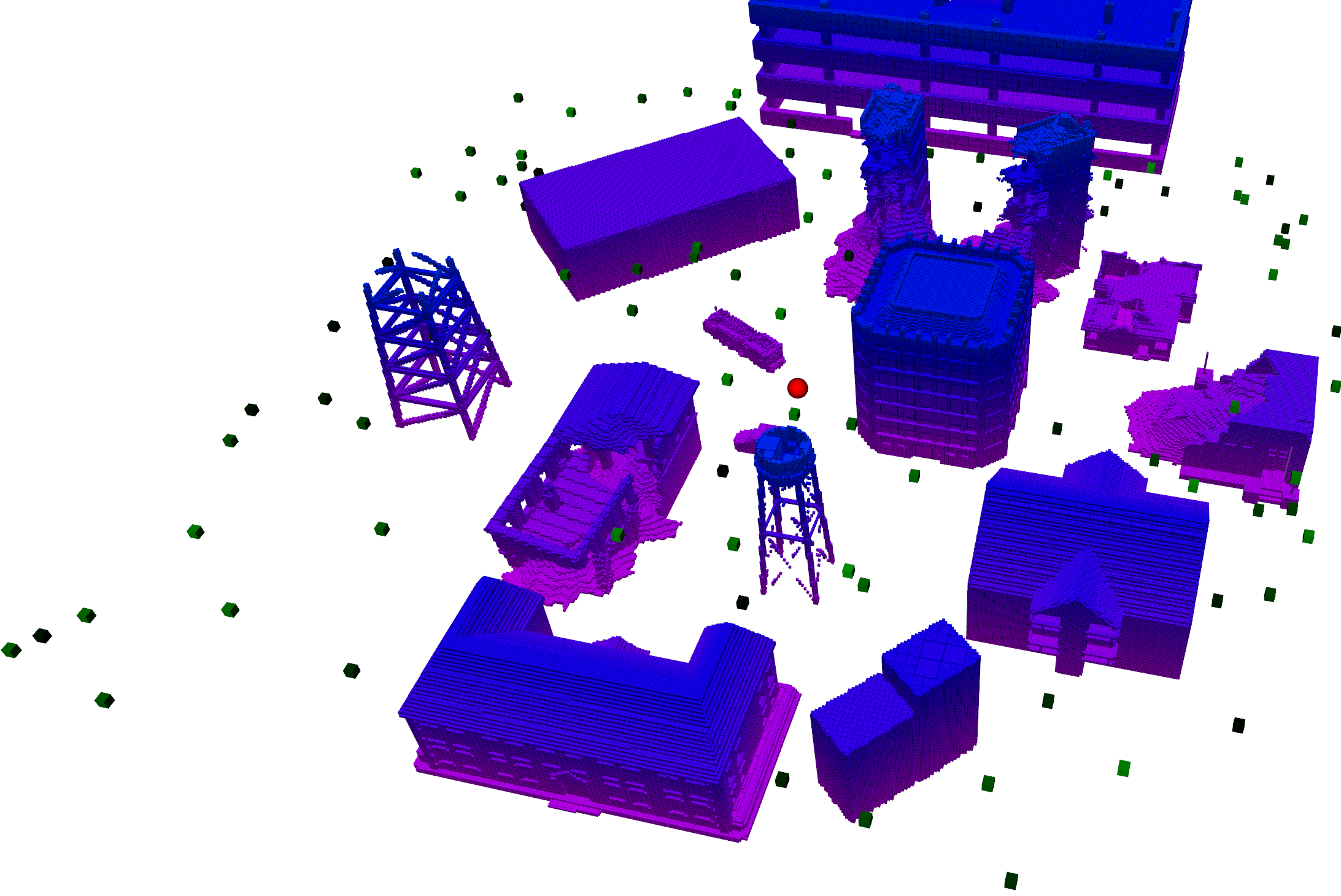}}
b)\hspace{0.2em}\mbox{\includegraphics[width=0.19\textwidth]{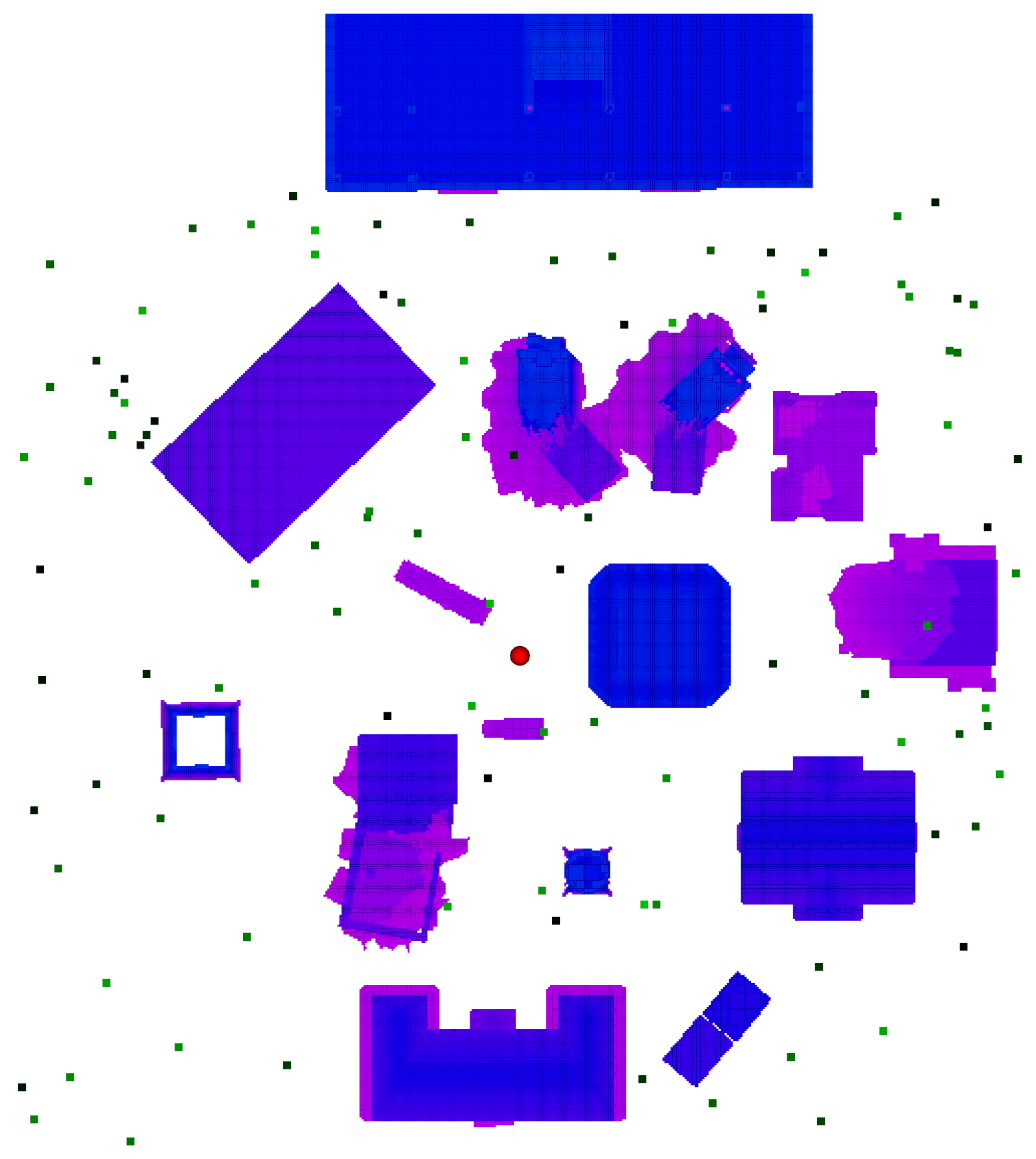}}
 \caption{The evaluation environment. a) 3D. b) Top-down. The start position is at the map center (red sphere). Goal positions are marked with squares which are brighter for larger height values.}
 \label{fig:arena}
\end{figure}

We use the following parameters for uniform planning as well as for $\textit{Level-1}$ of multiresolutional planning:
\begin{center}
\begin{tabular}{c|c|c|c|c}
$\rho$ & $\tau$ & $v_\text{max}$ & $u_\text{max}$ & $du$\\ \hline
16 & \SI{0.5}{\second} & \SI{4}{\meter\per\second} &  \SI{2}{\metre\per\square\second}  &  \SI{2}{\metre\per\square\second}
\end{tabular} .
\end{center}
The control set $\mathcal U^{(1)}$ for each dimension is obtained by discretizing $[-u_\text{max}, u_\text{max}]$ uniformly with resolution $du$. 
This results in \textit{Level-1} resolutions of $\Delta_p^{(1)} = \SI{0.25}{\meter}$ and $\Delta_v^{(1)} = \SI{1}{\meter\per\second}$.
The time cost factor $\rho=4u_\text{max}^2$ was chosen as proposed by Liu \etal~\cite{liu2018search}.

We evaluate the proposed methods in a simulated outdoor environment of size \SI{128}{}$\times$\SI{128}{\meter} containing several buildings (see~\reffig{fig:arena}).
The allowed flight altitude is \SIrange{0}{10}{\meter} and positions which are closer to obstacles than $\SI{1.5}{\meter}$ are considered invalid.

All planning methods are applied to the same set of \num{100} trajectory generation tasks.
The start position for all tasks is located at the map center with altitude \SI{2}{\meter}. 
Goal positions are sampled uniformly at random with the 
additional 
constraint that the distance to the closest obstacle or the map border is at least the spatial resolution of the corresponding level.

We are interested in evaluating the ability for frequent replanning.
Thus, after a trajectory is generated, we refine it by replanning from the MAV state that will be reached after one second.
This process is repeated until the goal state is reached.
For each planning task, we consider the maximum replanning time and expansion number over all replanning steps.
Furthermore, we report the costs of the fully refined trajectories.
If no solution is found within three million node expansions, we abort the planning process.
If a solution is found, the average number of replanning steps lies between $13$ and $18$.
Depending on the success rate, each method is thus applied to $900$ to $1400$ different planning tasks.

\subsection{1D Heuristic}
\begin{figure}
\begin{center}
a)\hspace{0.2em}\mbox{\includegraphics[width=0.2\textwidth]{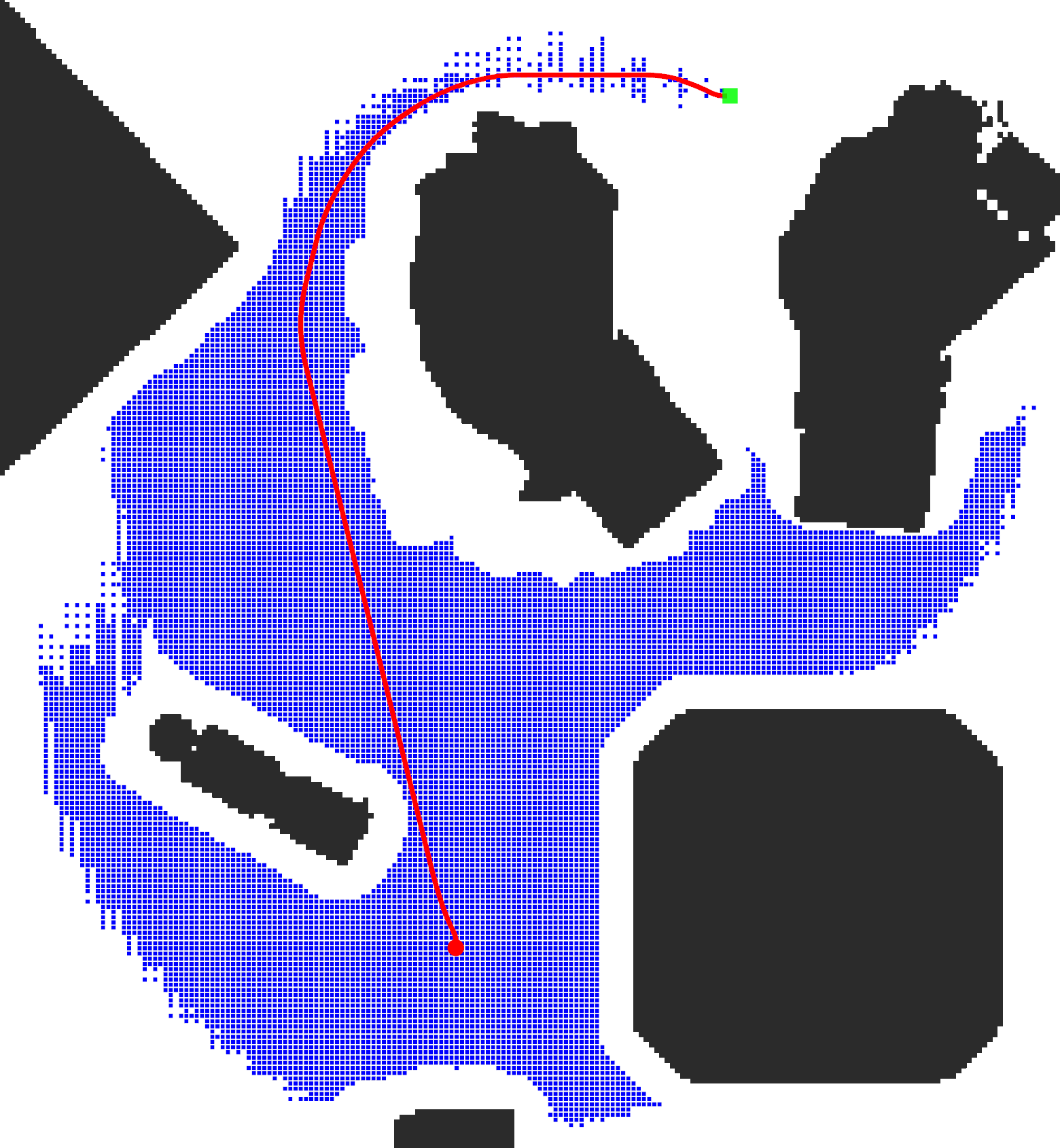}}
  \hspace{0.2em}
b)\hspace{0.2em}\mbox{\includegraphics[width=0.2\textwidth]{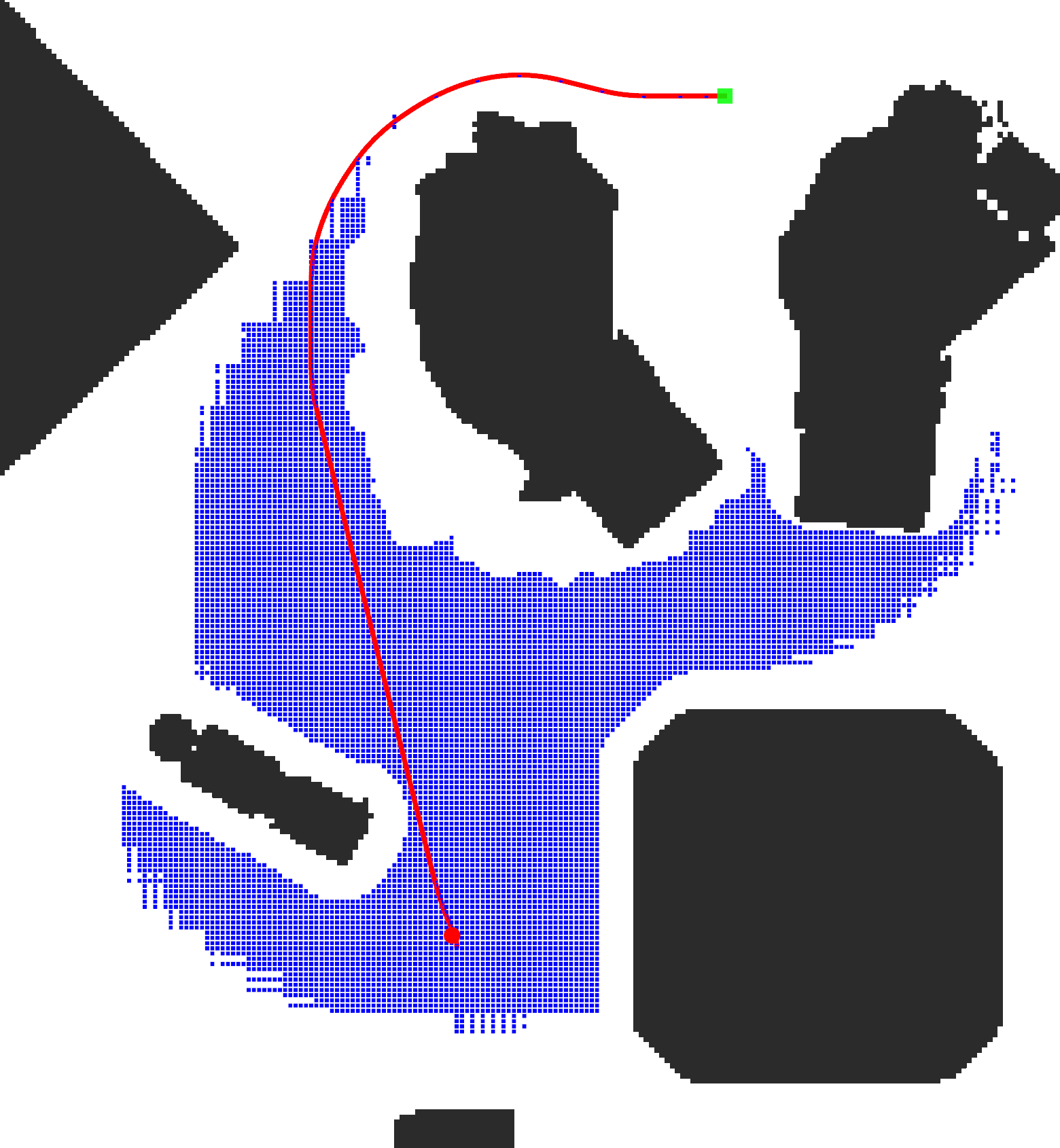}}
 \end{center}
 \caption{Expanded nodes using different heuristics for a planning task in two spatial dimensions. a) Baseline Heuristic (\num{115988} expansions). b) 1D Heuristic (\num{63769} expansions). Positions of expanded nodes are marked with blue squares. The start position is marked with a red circle, the goal with a green square. }
 \label{fig:heuristic}
\end{figure}

We evaluate the effect of our proposed 1D Heuristic separately for the different planning methods (see~\reftab{tab:heuristic}).
The number of node expansions is significantly reduced for all three methods.
The largest effect is achieved for uniform planning, where the number of expansions is reduced by $93\%$.
For the multiresolutional planning methods, there is a reduction of $80\%$ and $85\%$, respectively.
Since the 1D heuristic is not consistent, the generated trajectories are not optimal anymore.
However, the trajectory costs only increase about $0.75\%$ for uniform planning and multiresolution with variable time steps.
The increase for fixed level-dependent time steps is slightly larger with $1.46\%$.
\reffig{fig:heuristic} visualizes the set of expanded nodes for a uniform planning task in 2D.
Since our 1D heuristic significantly accelerates the A* search, we use it for all subsequent experiments.

\begin{table}
\newcommand{\mc}[3]{\multicolumn{#1}{#2}{#3}}
\caption{ Comparison of the baseline heuristic $h_\text{base}$ from~\cite{liu2017search} and our proposed 1D heuristic $h_\text{1D}$. For each planning method, the number of expansions and trajectory costs are averaged over all tasks where the corresponding method found valid trajectories with both heuristics. Note that the considered tasks might be different for different planning methods.}
\begin{center}
\begin{tabular}{rl|rrr}
& & Uniform & MRes\textsubscript{fixed} & MRes\textsubscript{variable} \\  \hline
 \multirow{2}{*}{$h_\text{base}$} & Expansions & \num{1133136} & \num{877041} & \num{890564}\\
 & Costs & \bf\num{243.69} & \bf\num{260.79} & \bf\num{270.14} \\  \hline
 \multirow{2}{*}{$h_\text{1D}$} & Expansions & \bf\num{75562} & \bf\num{175528} & \bf\num{135571}\\
 & Costs & \num{245.49} & \num{264.59} & \num{272.16}
\end{tabular}
\end{center}
\label{tab:heuristic}
 \vspace{-1em}
\end{table}

\subsection{Multiresolutional State Lattices with Standard A*}
\label{sec:eval_astar}
\begin{table}[t]
\caption{ Planning statistics for different state lattice representations. Maximum replanning times, number of expansions and trajectory costs are averaged over the tasks where all three planning methods found valid solutions. Additionally, the fraction of tasks for which the longest replanning step exceeds \SI{1}{\second} are given.}
\begin{center}
\begin{tabular}{l|rrrr}
 & Mean Maximum & Time & \multirow{2}{*}{Expansions} & \multirow{2}{*}{Costs}\\
 &  Replanning Time & > \SI{1}{\second} &  & \\ \hline 
Uniform & \SI{3.54}{\second} & $\mathbf{46.07}$\textbf{\%} & \num{228805} & \bf\num{261.2} \\
MRes\textsubscript{fixed} & \SI{6.44}{\second} & $62.92\%$ & \num{309526} & \num{273.2} \\
MRes\textsubscript{variable} & \bf\SI{2.73}{\second} & $58.43\%$ & \bf\num{160032} & \num{274.6}
\end{tabular}
\end{center}
\label{tab:astar}
\end{table}

To evaluate the effect of multiresolutional state lattices, we record for each trajectory generation task the maximal replanning time and maximum number of node expansions. 
\reftab{tab:astar} reports the averages of these values over the tasks where all three planning methods generated valid trajectories.
Interestingly, using multiresolutional state lattices with fixed level-dependent motion primitive durations significantly increases the maximum replanning time.
Furthermore, maximum replanning times exceed \SI{1}{\second} more often for both multiresolutional state lattice variants.
Although the size of the state space is reduced when using multiresolution, those methods tend to expand more states during A* search for some tasks.

\begin{table}[t]
\caption{ Planning statistics using the level-based expansion scheme. Maximum replanning times, number of expansions and trajectory costs are averaged over the same tasks as in~\reftab{tab:astar}.}
\begin{center}
\begin{tabular}{l|rrrr}
 & Mean Maximum & Time & \multirow{2}{*}{Expansions} & \multirow{2}{*}{Costs}\\
 &  Replanning Time & > \SI{1}{\second} &  & \\ \hline 
Uniform & \SI{0.65}{\second} & $16.85\%$ & \num{50880} & \num{273.8} \\
MRes\textsubscript{fixed} & \SI{0.57}{\second} & $12.36\%$ & \num{26078} & \bf\num{272.3} \\
MRes\textsubscript{variable} & \bf\SI{0.30}{\second} & $\mathbf{6.74}\textbf{\%}$ & \bf\num{17772} & \num{276.2}
\end{tabular}
\vspace*{-1.5em}
\end{center}
\label{tab:level_astar}
\end{table}

\begin{figure}[t]
a)\hspace*{-0.8em}\mbox{\includegraphics[width=0.24\textwidth]{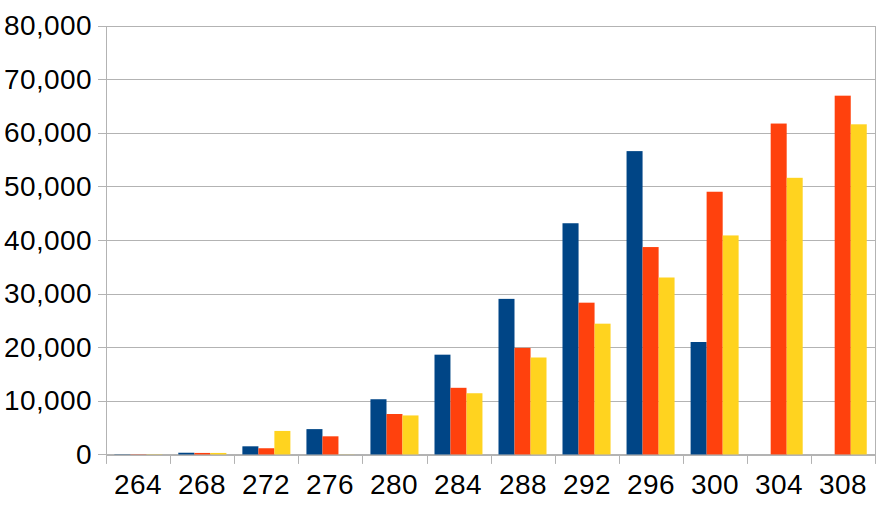}}
b)\hspace*{-0.8em}\mbox{\includegraphics[width=0.24\textwidth]{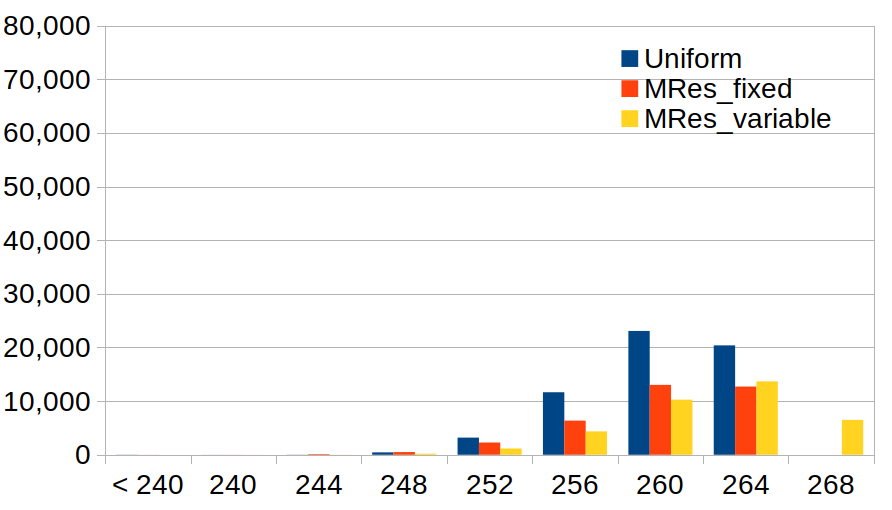}}
 \caption{Histograms for the $f$-values of each expanded state with different time cost weights $\rho$. a) $\rho=16$. b) $\rho=8$.}
 \label{fig:f_value_histogram}
 \vspace*{-0.5em}
\end{figure}

\begin{table*}[t]
\caption{ Comparison of the baseline~\cite{liu2017search} against the methods presented in this paper. Success denotes the fraction of tasks for which a valid solution was found within at most three million expansions. The number of tasks for which the longest replanning step exceeds \SI{1}{\second} includes tasks for which the corresponding method did not find a valid solution although one exists. Maximum replanning times, number of expansions and trajectory costs are averaged over the tasks where all methods found a valid solution. To compare the last three methods (using Level-A* and the 1D heuristic) against each other, please refer to~\reftab{tab:level_astar}, which also considers more challenging tasks not solved by the baseline.}
\begin{center}
 \vspace{-0.5em}
\begin{tabular}{l|rrrrr}
 &  \multirow{2}{*}{Success} & Mean Maximum &  \multirow{2}{*}{Time > \SI{1}{\second}} &  \multirow{2}{*}{Expansions} &  \multirow{2}{*}{Costs} \\
 &  & Replanning Time &  &  &  \\ \hline
Uniform, $h_\text{base}$, A*                     & $71.43\%$ & \SI{22.52}{\second} & $89.80\%$ & \num{1085740} & \bf\num{241.31} \\
MRes\textsubscript{fixed}, $h_\text{base}$, A*       & $79.59\%$ & \SI{17.87}{\second} & $89.80\%$ & \num{744640} & \num{249.91} \\
MRes\textsubscript{variable}, $h_\text{base}$, A*       & $87.76\%$ & \SI{14.21}{\second} & $87.76\%$ & \num{664036} & \num{253.85} \\ \hline
Uniform, $h_\text{1D}$, A*                       & $91.84\%$ & \SI{1.14}{\second} & $51.02\%$ & \num{70085} & \num{243.16} \\
MRes\textsubscript{fixed}, $h_\text{1D}$, A*         & $95.92\%$ & \SI{2.87}{\second} & $66.33\%$ & \num{143822} & \num{254.33} \\
MRes\textsubscript{variable}, $h_\text{1D}$, A*         & $98.98\%$ & \SI{1.60}{\second} & $62.24\%$ & \num{89085} & \num{255.85} \\ \hline
Uniform, $h_\text{base}$, Level-A*               & $95.92\%$ & \SI{0.41}{\second} & $29.59\%$ & \num{24190} & \num{246.15} \\
MRes\textsubscript{fixed}, $h_\text{base}$, Level-A* & $97.96\%$ & \SI{0.20}{\second} & $20.41\%$ & \bf\num{5684} & \num{246.69} \\
MRes\textsubscript{variable}, $h_\text{base}$, Level-A* & $\mathbf{100}$\textbf{\%} & \bf\SI{0.19}{\second} & $16.33\%$ & \num{8505} & \num{248.84} \\ \hline
Uniform, $h_\text{1D}$, Level-A*                       & $\mathbf{100}$\textbf{\%} & \SI{0.20}{\second} & $22.45\%$ & \num{16739} & \num{253.94} \\
MRes\textsubscript{fixed}, $h_\text{1D}$, Level-A*   & $\mathbf{100}$\textbf{\%} & \SI{0.21}{\second} & $20.41\%$ & \num{8284} & \num{252.15} \\
MRes\textsubscript{variable}, $h_\text{1D}$, Level-A*   & $\mathbf{100}$\textbf{\%} & \SI{0.20}{\second} & $\mathbf{12.24}$\textbf{\%} & \num{10945} & \num{255.28}
\end{tabular}
\end{center}
\label{tab:overview}
 \vspace{-1.5em}
\end{table*}

To further investigate this behavior, we have a closer look at the tasks where uniform planning outperforms multi\-resolution planning.
We compare the $f$-values of each state $s$, \ie the sum of the costs for moving from the start state to $s$ and the heuristic costs for reaching the goal from $s$.
\reffig{fig:f_value_histogram} shows histograms for the $f$-values of each expanded state during an example search for the different state lattice representations.
The number of states grows approximately cubic with increasing $f$-value.
Due to the reduced number of states, the curves for the multiresolutional lattices grow at lower rates.
The A* algorithm expands all states whose $f$-value are smaller than the optimal path costs.
Since the costs of the paths in multiresolutional lattices are higher than the optimal costs in the uniform lattice, states up to a higher $f$-value have to be expanded.
Due to the cubic increase, this overhead might result in an overall larger amount of expanded states for multiresolutional search.
Note that the difference between the maximum $f$-values for the different state representations is only $8$.
This corresponds to the smallest possible cost for one action.
When reducing the time cost factor $\rho$, multiresolutional planning expands fewer states than uniform planning (\reffig{fig:f_value_histogram}\,b), but the resulting trajectory durations might be suboptimal.
Instead, the issue can be addressed by removing the constraint that all states whose $f$-value are smaller than the optimal path costs have to be expanded.
This is evaluated in the next section.

\subsection{Level-Based Expansion Scheme}
We apply the level-based expansion scheme from~\refsec{sec:level_astar} to all three state lattice representations and evaluate them on the same set of tasks that was used in~\refsec{sec:eval_astar}.
The results are summarized in~\reftab{tab:level_astar}.
All three methods benefit from using the adapted expansion scheme.
While the average maximum replanning times for uniform state lattices are reduced by around $80\%$, the reduction for multiresolutional lattices is even higher with around $90\%$.
Similar reductions are achieved with respect to the number of state expansions.
The trajectory costs for uniform lattices increases by roughly $5\%$.
Interestingly, the costs for multiresolutional planning do not change much.
When using fixed level-dependent time steps, the costs even decrease slightly, which is possible because the used heuristic is not consistent.

In~\reftab{tab:overview}, we compare the method of~\cite{liu2017search} against all methods presented in this work.
Note that average planning times, expansion numbers and trajectory costs only consider tasks solved by the baseline.
The costs of all trajectories are at most $6\%$ longer than the optimal trajectory generated by the baseline.
However, the average maximum replanning time can be reduced by up to two orders of magnitude.

\subsection{Online Replanning}
\begin{figure}
\begin{center}
 \includegraphics[width=0.4\textwidth]{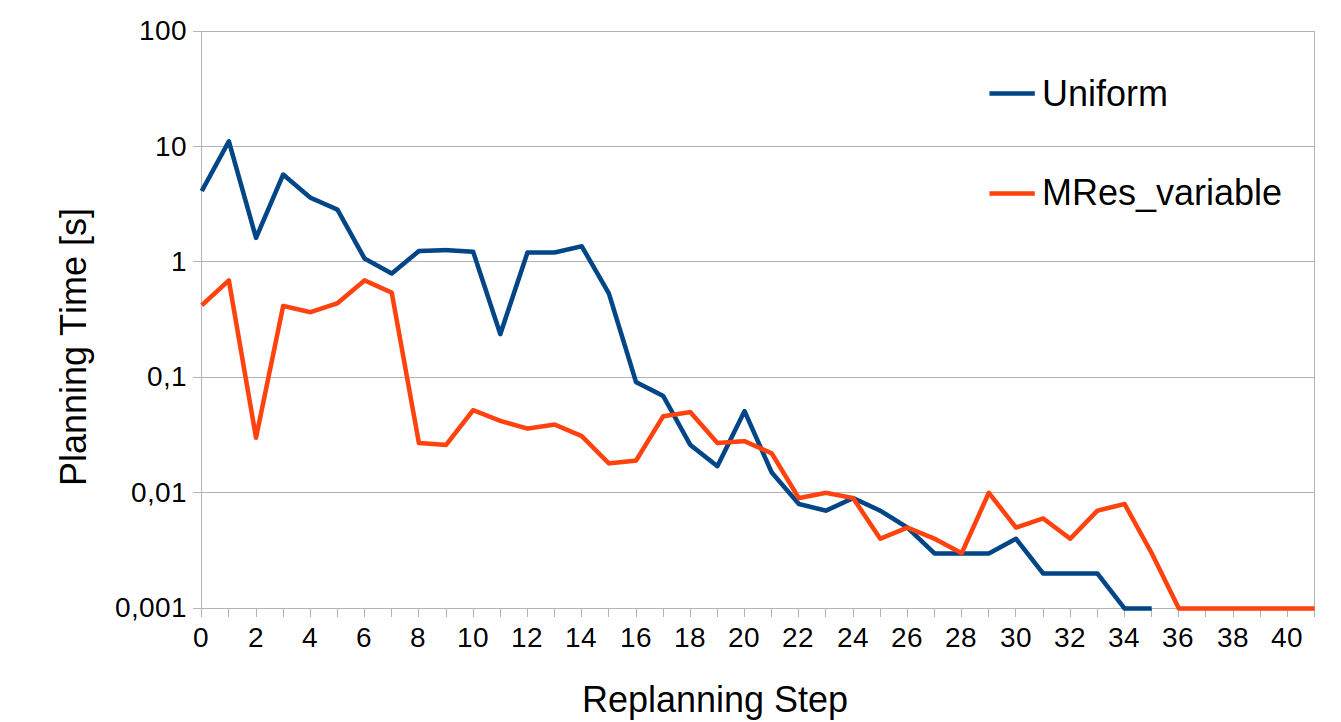}
\end{center}
 \caption{Replanning times in simulation.}
 \label{fig:online_replanning}
 \vspace{-1.5em}
\end{figure}
\begin{figure}
\begin{center}
 \includegraphics[width=0.4\textwidth]{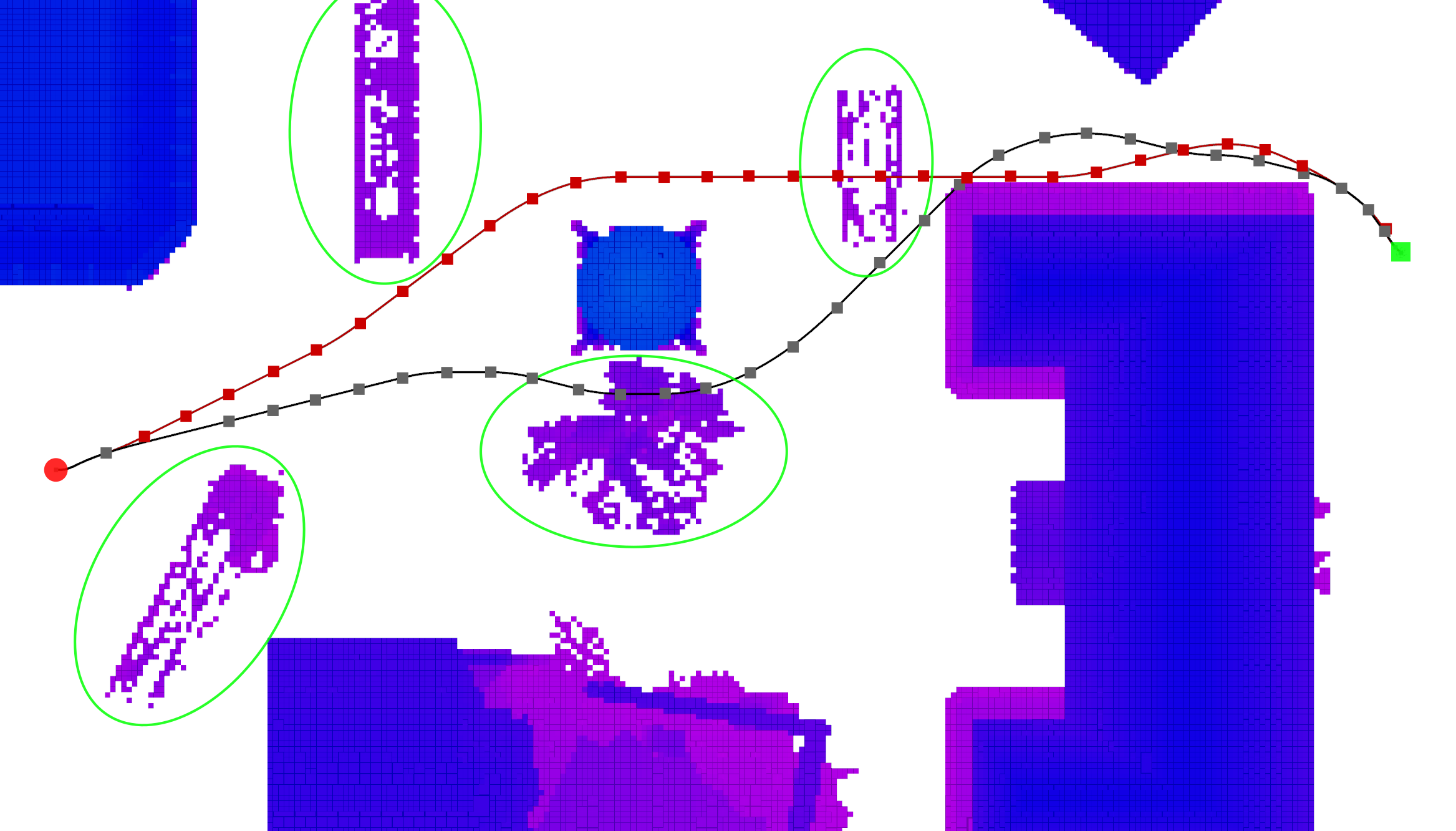}
\end{center}
 \caption{Trajectories from online replanning (top-down view). Obstacle heights are represented by the color, reaching from purple (low) to blue (high). The trajectory of MRes\textsubscript{variable} is shown in red, the trajectory from uniform planning in black. The start position is marked by the red circle and the goal by a green square. Start positions of replanning steps are marked with red and gray squares. The initially unknown obstacles are circled.}
 \label{fig:trajectories}
 \vspace{-1.0em}
\end{figure}

Finally, we test our approach in simulation using the RotorS simulator~\cite{furrer2016rotors}.
We start with an initial OctoMap~\cite{hornung13auro} of the environment (\reffig{fig:arena}), but add additional unmapped static obstacles such that frequent replanning is necessary.
Replanning is triggered at \SI{1}{\hertz} and the map is constantly updated using measurements of a simulated 3D laser scanner.
\reffig{fig:online_replanning} shows the corresponding planning times of uniform lattices and MRes\textsubscript{variable}.
Both methods use our 1D heuristic and the level-based expansion scheme.
While our approach has a maximal replanning time of \SI{0.69}{\second}, uniform planning exceeds \SI{1}{\second} in one third of all replanning steps.
\reffig{fig:trajectories} shows the corresponding trajectories.
A 3D view of the environment and the trajectory generated by our method is given in~\reffig{fig:teaser}.

Additionally, we successfully employed our approach to online trajectory planning on a real MAV.
For details on the integrated system and corresponding experiments, we refer to~\cite{schleich2021icuas}.

\section{Conclusion}
\label{sec:conclusion}
In this paper, we introduced high-dimensional local multiresolution state lattices.
MAV velocities are directly incorporated into the planning to generate dynamically feasible trajectories.
We showed that multiresolution combined with standard A* search might result in higher planning times compared to uniform state lattices.
However, in combination with a level-based expansion scheme, multiresolutional state lattices significantly reduce the maximal planning times, while only moderately increasing trajectory costs.
For some challenging tasks, planning times still exceed \SI{1}{\second}, but our approach was able to maintain a replanning frequency of \SI{1}{\hertz} for most cases.
In summary, we demonstrated how multiresolution can increase the applicability of search-based high-dimensional trajectory planning for large dynamic environments.
 
\IEEEtriggeratref{12}
\bibliographystyle{IEEEtranBST/IEEEtran}
\bibliography{literature_references}

\begin{thebibliography}{10}
\providecommand{\url}[1]{#1}
\csname url@rmstyle\endcsname
\providecommand{\newblock}{\relax}
\providecommand{\bibinfo}[2]{#2}
\providecommand\BIBentrySTDinterwordspacing{\spaceskip=0pt\relax}
\providecommand\BIBentryALTinterwordstretchfactor{4}
\providecommand\BIBentryALTinterwordspacing{\spaceskip=\fontdimen2\font plus
\BIBentryALTinterwordstretchfactor\fontdimen3\font minus
  \fontdimen4\font\relax}
\providecommand\BIBforeignlanguage[2]{{%
\expandafter\ifx\csname l@#1\endcsname\relax
\typeout{** WARNING: IEEEtran.bst: No hyphenation pattern has been}%
\typeout{** loaded for the language `#1'. Using the pattern for}%
\typeout{** the default language instead.}%
\else
\language=\csname l@#1\endcsname
\fi
#2}}

\bibitem{hart1968formal}
P.~E. Hart, N.~J. Nilsson, and B.~Raphael, ``A formal basis for the heuristic
  determination of minimum cost paths,'' \emph{{IEEE T}ransactions on Systems
  Science and Cybernetics}, vol.~4, no.~2, pp. 100--107, 1968.

\bibitem{lavalle1998rapidly}
S.~M. LaValle, ``Rapidly-exploring random trees: {A} new tool for path
  planning,'' \emph{Computer Science Dept., Iowa State University}, 1998.

\bibitem{richter2016polynomial}
C.~Richter, A.~Bry, and N.~Roy, ``Polynomial trajectory planning for aggressive
  quadrotor flight in dense indoor environments,'' in \emph{Robotics
  Research}.\hskip 1em plus 0.5em minus 0.4em\relax Springer, 2016, pp.
  649--666.

\bibitem{koyuncu2008probabilistic}
E.~Koyuncu and G.~Inalhan, ``A probabilistic {B}-spline motion planning
  algorithm for unmanned helicopters flying in dense {3D} environments,'' in
  \emph{IEEE/RSJ International Conference on Intelligent Robots and Systems
  (IROS)}, 2008, pp. 815--821.

\bibitem{kalakrishnan2011stomp}
M.~Kalakrishnan, S.~Chitta, E.~Theodorou, P.~Pastor, and S.~Schaal, ``{STOMP:
  S}tochastic trajectory optimization for motion planning,'' in \emph{IEEE
  International Conference on Robotics and Automation (ICRA)}, 2011, pp.
  4569--4574.

\bibitem{nieuwenhuisen2019search}
M.~Nieuwenhuisen and S.~Behnke, ``Search-based {3D} planning and trajectory
  optimization for safe micro aerial vehicle flight under sensor visibility
  constraints,'' in \emph{IEEE International Conference on Robotics and
  Automation (ICRA)}, 2019, pp. 9123--9129.

\bibitem{liu2017search}
S.~Liu, N.~Atanasov, K.~Mohta, and V.~Kumar, ``Search-based motion planning for
  quadrotors using linear quadratic minimum time control,'' in \emph{IEEE/RSJ
  International Conference on Intelligent Robots and Systems (IROS)}, 2017, pp.
  2872--2879.

\bibitem{liu2018search}
S.~Liu, K.~Mohta, N.~Atanasov, and V.~Kumar, ``Search-based motion planning for
  aggressive flight in {SE(3)},'' \emph{IEEE Robotics and Automation Letters},
  vol.~3, no.~3, pp. 2439--2446, 2018.

\bibitem{behnke2003local}
S.~Behnke, ``Local multiresolution path planning,'' in \emph{Robot Soccer World
  Cup}.\hskip 1em plus 0.5em minus 0.4em\relax Springer, 2003, pp. 332--343.

\bibitem{nieuwenhuisen2014hierarchical}
M.~Nieuwenhuisen and S.~Behnke, ``Hierarchical planning with {3D} local
  multiresolution obstacle avoidance for micro aerial vehicles,'' in
  \emph{Joint 45th International Symposium on Robotics (ISR) and 8th German
  Conference on Robotics (ROBOTIK)}.\hskip 1em plus 0.5em minus 0.4em\relax
  VDE, 2014.

\bibitem{nieuwenhuisen2016local}
------, ``Local multiresolution trajectory optimization for micro aerial
  vehicles employing continuous curvature transitions,'' in \emph{IEEE/RSJ
  International Conference on Intelligent Robots and Systems (IROS)}, 2016, pp.
  3219--3224.

\bibitem{du2020multi}
W.~Du, F.~Islam, and M.~Likhachev, ``{Multi-Resolution A*},'' in
  \emph{International Symposium on Combinatorial Search (SoCS)}, 2020.

\bibitem{gonzalez2016adaptive}
A.~Gonz{\'a}lez-Sieira, M.~Mucientes, and A.~Bugar{\'\i}n, ``An adaptive
  multi-resolution state lattice approach for motion planning with
  uncertainty,'' in \emph{Robot 2015: Second Iberian Robotics
  Conference}.\hskip 1em plus 0.5em minus 0.4em\relax Springer, 2016, pp.
  257--268.

\bibitem{gonzalez2019graduated}
------, ``Graduated fidelity lattices for motion planning under uncertainty,''
  in \emph{IEEE International Conference on Robotics and Automation (ICRA)},
  2019, pp. 5908--5914.

\bibitem{pivtoraiko2008differentially}
M.~Pivtoraiko and A.~Kelly, ``Differentially constrained motion replanning
  using state lattices with graduated fidelity,'' in \emph{IEEE/RSJ
  International Conference on Intelligent Robots and Systems (IROS)}, 2008, pp.
  2611--2616.

\bibitem{andersson2018receding}
O.~Andersson, O.~Ljungqvist, M.~Tiger, D.~Axehill, and F.~Heintz,
  ``Receding-horizon lattice-based motion planning with dynamic obstacle
  avoidance,'' in \emph{IEEE Conference on Decision and Control (CDC)}, 2018,
  pp. 4467--4474.

\bibitem{likhachev2009planning}
M.~Likhachev and D.~Ferguson, ``Planning long dynamically feasible maneuvers
  for autonomous vehicles,'' \emph{The International Journal of Robotics
  Research}, vol.~28, no.~8, pp. 933--945, 2009.

\bibitem{rufli2009smooth}
M.~Rufli, D.~Ferguson, and R.~Siegwart, ``Smooth path planning in constrained
  environments,'' in \emph{IEEE International Conference on Robotics and
  Automation (ICRA)}, 2009, pp. 3780--3785.

\bibitem{petereit2013mobile}
J.~Petereit, T.~Emter, and C.~W. Frey, ``Mobile robot motion planning in
  multi-resolution lattices with hybrid dimensionality,'' \emph{IFAC
  Proceedings Volumes}, vol.~46, no.~10, pp. 158--163, 2013.

\bibitem{furrer2016rotors}
F.~Furrer, M.~Burri, M.~Achtelik, and R.~Siegwart, ``{RotorS—A} modular
  {G}azebo {MAV} simulator framework,'' in \emph{Robot Operating System
  (ROS)}.\hskip 1em plus 0.5em minus 0.4em\relax Springer, 2016, pp. 595--625.

\bibitem{hornung13auro}
\BIBentryALTinterwordspacing
A.~Hornung, K.~M. Wurm, M.~Bennewitz, C.~Stachniss, and W.~Burgard,
  ``{OctoMap}: An efficient probabilistic {3D} mapping framework based on
  octrees,'' \emph{Autonomous Robots}, 2013. [Online]. Available:
  \url{http://octomap.github.com}
\BIBentrySTDinterwordspacing

\bibitem{schleich2021icuas}
D.~Schleich, M.~Beul, J.~Quenzel, and S.~Behnke, ``Autonomous flight in unknown
  {GNSS}-denied environments for disaster examination,'' \emph{arXiv preprint
  arXiv:2103.11742}, 2021.

\end{thebibliography}

\end{document}